\documentclass{bmvc2k}
\usepackage{epsfig}
\usepackage{graphicx}
\usepackage{amsmath}
\usepackage{amssymb}

\usepackage{url}            % simple URL typesetting
\usepackage{booktabs}       % professional-quality tables
\usepackage{amsfonts}       % blackboard math symbols
\usepackage{nicefrac}       % compact symbols for 1/2, etc.
\usepackage{microtype}      % microtypography

\usepackage{capt-of}
\usepackage{comment}
\usepackage{color}
\usepackage{pifont}
\usepackage{soul}

\usepackage{array}
\usepackage{cases}
\usepackage{booktabs}
\usepackage{multirow}
\usepackage{colortbl}
\usepackage[linesnumbered, ruled]{algorithm2e}
\usepackage{algpseudocode}
\newcommand{\etal}{\textit{et al.}}
\newcommand{\eg}{\textit{e.g.}}
\newcommand{\ie}{\textit{i.e.}}
\newcommand{\eq}{Eq.}

\newcommand{\etc}{\textit{etc.}}

\newcommand{\proposed}{SCS-UIT}
\newcommand{\uit}{UIT}                      % Image-to-image Translation 
\definecolor{mygray}{gray}{.9}

\def  \LL      {\mathcal{L}}                 % Loss function
\def  \XA      {\mathcal{X}_A}				 % Image Domain A
\def  \XB      {\mathcal{X}_B}				 % Image Domain B
\def  \SA      {\mathcal{S}_A}				 % Style Domain A
\def  \SB      {\mathcal{S}_B}				 % Style Domain B
\def  \C       {\mathcal{C}}				 % Shared Content Domain
\def  \H       {\mathcal{H}}				 % High-level task Domain
\def  \DIT     {DI-HV}					 	 % Domain-invariant high-level vision
\title{Separating Content and Style for Unsupervised Image-to-Image Translation}
\addauthor{Yunfei Liu}{lyunfei@buaa.edu.cn}{1}
\addauthor{Haofei Wang}{wanghf@pcl.ac.cn}{2}
\addauthor{Yang Yue}{yuey23@buaa.edu.cn}{3}
\addauthorv1{Feng Lu}{lufeng@buaa.edu.cn}{1}{2, \ding{41}}

\addinstitution{
 State Key Laboratory of Virtual Reality Technology and Systems, School of Computer Science and Engineering, \\
 Beihang University, Beijing, China\\
}
\addinstitution{
 Peng Cheng Laboratory.\\
 Shenzhen, China
}
\addinstitution{
 Beihang University, Beijing, China\\
}

\runninghead{Liu, Wang, Yue, Lu}{\proposed}

\def\eg{\emph{e.g}\bmvaOneDot}

\def\etal{\emph{et al}\bmvaOneDot}

\begin{document}

\maketitle

\begin{abstract}
Unsupervised image-to-image translation aims to learn the mapping between two visual domains with unpaired samples. Existing works focus on disentangling domain-invariant content code and domain-specific style code individually for multimodal purposes. However, less attention has been paid to interpreting and manipulating the translated image. In this paper, we propose to separate the content code and style code simultaneously in a unified framework. Based on the correlation between the latent features and the high-level domain-invariant tasks, the proposed framework demonstrates superior performance in multimodal translation, interpretability and manipulation of the translated image. Experimental results show that the proposed approach outperforms the existing unsupervised image translation methods in terms of visual quality and diversity. Code and data have been released at \url{https://github.com/DreamtaleCore/SCS-UIT}.
\end{abstract}

%-------------------------------------------------------------------------
\section{Introduction}
Unsupervised image-to-image translation (\uit) attracts great attention since it can learn the mapping between different visual domains without paired data. Numerous computer vision and graphics problems can be formulated as \uit~problems, such as image super-resolution~\cite{A:dong2015image,A:ledig2017photo,A:wang2018esrgan}, in-painting~\cite{A:pathak2016context}, style-transfer~\cite{A:johnson2016perceptual}, and other low-level vision tasks~\cite{A:liu2020usi3d}. 

Learning the mappings between two visual domains are inherently multimodal, \ie, a single input may correspond to multiple possible outputs.
For the source visual domain $\XA$ and target domain $\XB$, given the source image $x_A \in \XA$,  \uit~ aims to keep the content of $x_A$, and to turn its style into the target domain. However, It is not easy to define the content and style precisely. As with the existing methods, we make similar assumptions~\cite{I:MUNIT,I:DRIT} for the content code and style code.
In general, the content is shared between two domains and often refers more to semantics, layouts, and spatial arrangements of the image, while the style involves more in color, tones, and textures, and other domain-specific features.

\begin{figure*} [htbp]
	\begin{center}
		\includegraphics[width=\linewidth]{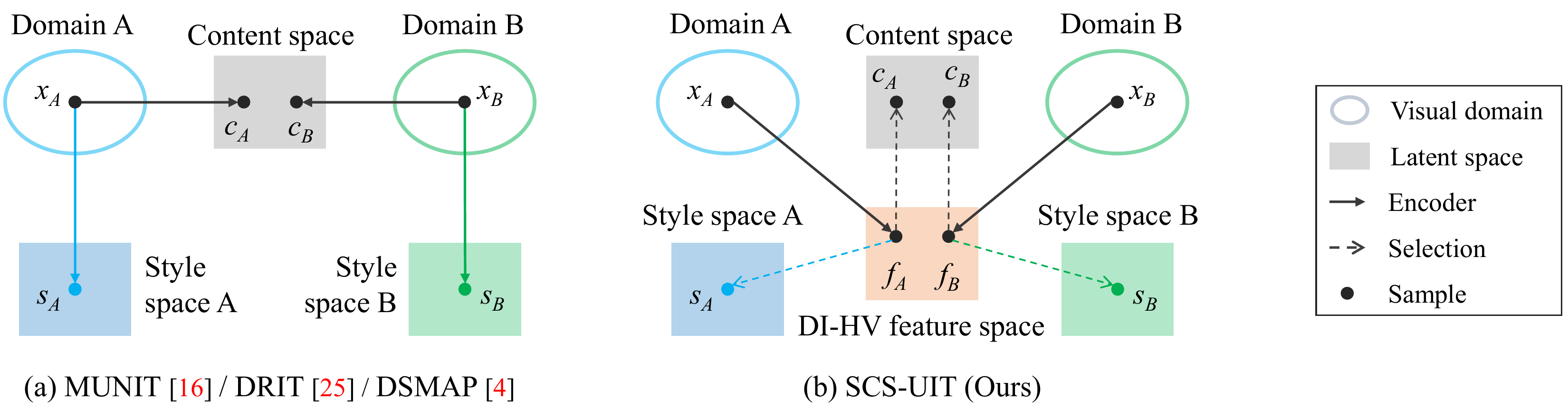} % set width=0.85 for final version
	\end{center}
	\vspace{-0.2cm}
	\caption{Comparison of the existing approaches on unsupervised image-to-image translation and our proposed method. (a) MUNIT~\cite{I:MUNIT}, DRIT~\cite{I:DRIT} and DSMAP~\cite{I:chang2020dsmap} extract the shared content code and the domain-specific style code through two independent encoders. (b) The proposed SCS-UIT framework separates the content code and style code in feature space on the basis of correlation to the domain-invariant high-level vision task.}
	\label{fig:difference}
	\vspace{-0.2cm}
\end{figure*}

To achieve such translation, several recent works using cycle consistency~\cite{I:cycleGAN,I:UNIT} and disentangled representations have been proposed.
For example, MUNIT~\cite{I:MUNIT} and DRIT~\cite{I:DRIT} encode an input image $x$ into two different features, the style feature $s$ and the content feature $c$, as illustrated in Fig.~\ref{fig:difference}(a). In their settings, domain $\XA$ and $\XB$ have their own domain-specific feature spaces for styles $\SA$ and $\SB$. Meanwhile, they share the same latent domain-invariant space $\C$ for the content. Thus, given a source image $x_A$ and a target image $x_B$, they are encoded as $(c_A, s_A)$ and $(c_B, s_B)$, respectively. The content codes $c_A$ and $c_B$ belong to $\C$, while style codes $s_A$ and $s_B$ belong to $\SA$ and $\SB$, respectively. For cross domain translation task $\XA \mapsto \XB$, the content code $c_A$ and the style code $s_B$ are combined together for synthesizing the result with content from $x_A$ and style from $x_B$.

Recently, several approaches have been proposed to learn the content code with additional high-level tasks under supervision. INIT~\cite{I:Shen2019Towards} uses an additional model to predict the bounding boxes of the instances in the image, and applies both instance-level style and global style to the target image spatially. TransGaGa~\cite{I:Wu2019TransGaGa} uses landmarks as the shared geometry information and translates images from two visual domains with apparent geometry gap. However, these works mainly focus on the translation between two domains, while the interpretation and manipulation during the translation have not been well-explored.

In this work, we proposed the Separating Content and Style for Unsupervised Image-to-image Translation (\proposed) framework, which extracts and separates content and style in a unified feature space. More specifically, we first define the Domain-Invariant High-level Vision task (\DIT) for \uit~task. Then we separate style code and content code from different feature channels on the basis of the correlation to \DIT. The proposed framework has several good properties: 1) it supports multimodal translation within a single encoder, 2) it achieves good interpretability and 3) the translated results can be easily manipulated. 
In summary, the primary contributions of this work are as follows:
\begin{itemize}
	\item We introduced the Domain-Invariant High-level Vision (\DIT) task for unsupervised image-to-image translation. Based on the correlation to \DIT, a new unsupervised image-to-image translation framework (\proposed) is proposed.
	\item The \proposed~not only eases the interpretation of the translation but also enables us to interactively edit the semantic regions in the translated image. 
	\item Experimental results demonstrate that our method outperforms the state-of-the-art methods on four image translation tasks.
\end{itemize}

%-------------------------------------------------------------------------
\section{Related Work}

\noindent\textbf{Image-to-image translation} aims at learning the mapping from the source domain to the target domain.
Pix2pix~\cite{I:isola2017Pix2pix} is the first work using conditional GANs for image-to-image translation. Following~\cite{I:isola2017Pix2pix}, several works seek to address other computer vision tasks, such as super-resolution~\cite{A:ledig2017photo,A:wang2018esrgan}, domain adaption~\cite{A:hoffman2017cycada,A:li2019learning},  colorization~\cite{A:bahng2018coloring,A:yoo2019coloring} and low-light enhancement~\cite{A:lv2018mbllen,A:lv2020integrated}.

However, paired data is not always available in practical applications. 
CycleGAN\cite{I:cycleGAN} and UNIT ~\cite{I:UNIT} are trained with unpaired data, which equips cycle consistency for UIT. DiscoGAN~\cite{I:kim2017DiscoGAN}, DualGAN~\cite{I:yi2017dualgan} and SelectionGAN~\cite{I:selectionGAN} are proposed following the idea of cycle-consistency.
Other works apply attention mechanism to UIT~\cite{I:emami2020spa,I:kim2019u,I:mejjati2018attention,I:yang2019show}.

Some works attempt to address one-to-many translation or many-to-many translation. By assuming that an image can be decomposed into a domain-invariant content code and a domain-specific style code, CIIT~\cite{I:lin2018CIIT}, EG-UNIT~\cite{I:ma2018exemplar}, MUNIT~\cite{I:MUNIT} and DRIT~\cite{I:DRIT} learn a one-to-many mapping (\ie multimodal) between the two image domains in an unsupervised settings. Other works add additional constrains to improve style-content disentanglement in image-to-image translation\cite{I:chang2020dsmap,I:gabbay2020stylecontent,I:na2019miso,I:romero2019smit,I:roy2020trigan,I:zheng2019generative}.
More recently, high-level vision tasks such as object landmark detection~\cite{I:Wu2019TransGaGa}, instance bound box detection~\cite{I:Shen2019Towards} and semantic parsing~\cite{I:roy2019semantics,I:wang2019controlling} are used for translation. However, they usually require an specifically designed architecture for these tasks.
Unlike the aforementioned methods, we propose a correlation-based framework to achieve multimodal UIT, which also enables us to interpret the translation process and manipulate the translation result.

\noindent\textbf{Interpreting the learned CNNs} helps us to gain more insights of the network design. There are two branches, one branch is to explain individual network decisions using informative feature maps of instances~\cite{V:qin2018convolutional}. Another branch is to explore the activation of units in CNNs using modified back-propagation~\cite{V:sundararajan2017axiomatic,V:Maaten08visualizingdata}. 
Morcos~\etal~\cite{V:morcos2018importance} examined the effect of individual units by an ablation study. Bau~\etal~\cite{V:bau2018GANdissection} visualized the GANs by manipulating specific channels in the latent space of the image. To manipulate the content in image translation, disentangling factors of image variations has attracted much attention~\cite{R:chen2018isolating,G:StyleGAN,G:StyleGAN2}. These works inspire us to interpret and manipulate the image-to-image translation results.

%-------------------------------------------------------------------------
\section{Methodology}

In this paper, we propose to separate content and style for unsupervised image-to-image translation (\proposed). This framework reveals the learned relationship of representation between content and style and separates them within one encoder. Therefore, we can interpret and manipulate the translated images semantically based on the learned model. 
Before diving into the design of the \proposed, we first give a definition of the domain-invariant high-level vision task, then give an overview of our method followed by the details.

\begin{figure}[htbp]
	\begin{minipage}[c]{0.52\textwidth}
		\includegraphics[width=\textwidth]{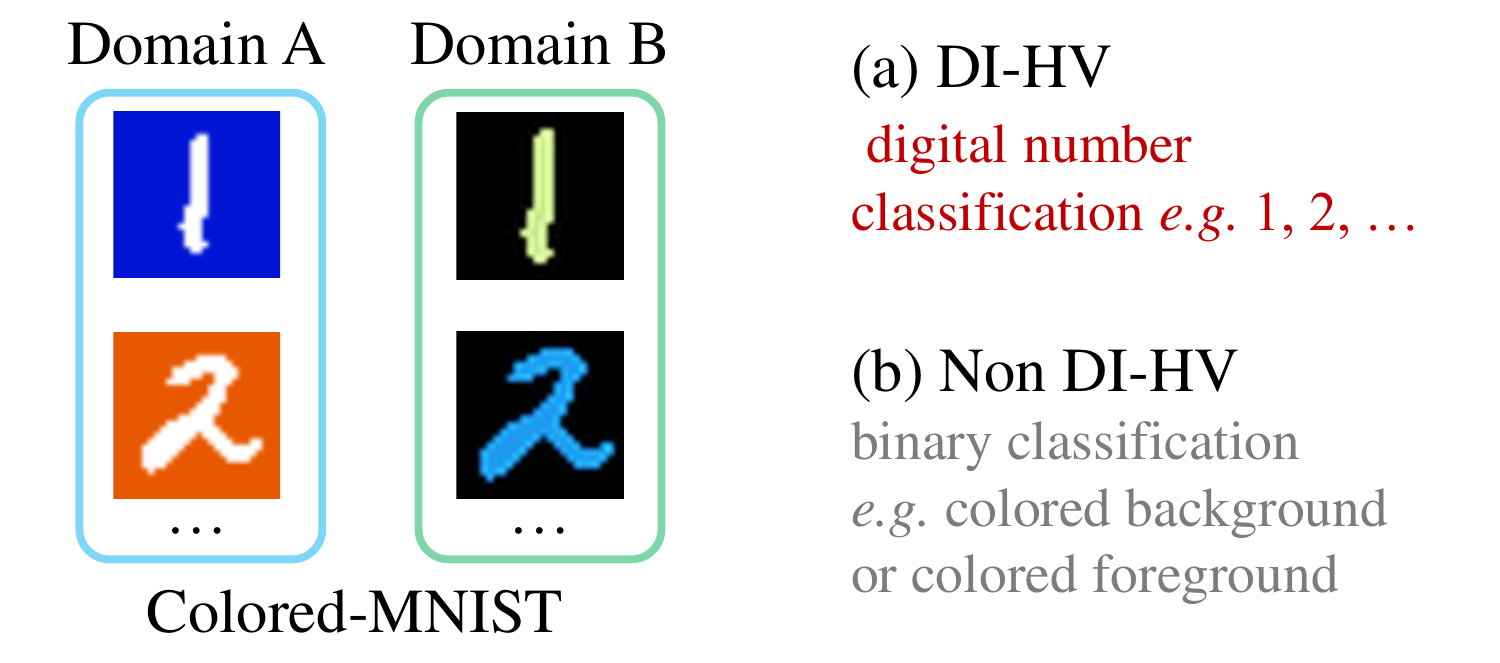}
	\end{minipage}\hfill
	\begin{minipage}[c]{0.47\textwidth}
		\caption{
			Comparisons between Domain-Invariant High-level Vision (\DIT) tasks and Non-\DIT~tasks on the Colored-MNIST dataset. (a) DI-HV are the tasks sharing between different visual domains. (b) Non \DIT~pay more attention to the domain-specific features.
		} \label{fig:DI_HVT}
	\end{minipage}
\vspace{-0.6cm}
\end{figure}

\subsection{Domain-invariant high-level vision task}

We denote $\mathcal{T}_A$ as the set of all the possible vision tasks that can be performed on $\XA$, and $\mathcal{T}_B$ as the set of all the tasks for $\XB$. Domain-invariant high-level vision task (\DIT)~is defined as $\mathcal{T}_A \cap \mathcal{T}_B$.
For an instance, Fig.~\ref{fig:DI_HVT} illustrates a typical example of \DIT. In this case, the \DIT~is the digital number classification task, as though the color tones are different between two domains, which can be taken as a non \DIT task. \DIT~can be easily extended to other image-to-image translations, \eg, face parsing could be a \DIT~for man $\leftrightarrows$ woman translation or cat face $\leftrightarrows$ human face, scene semantic segmentation could be a \DIT~for season translation or CG image $\leftrightarrows$ real image, number classification for ColoredMNSIT, \etc.

\subsection{Method overview}

\begin{figure}[htbp]
	\begin{center}
		\includegraphics[width=\linewidth]{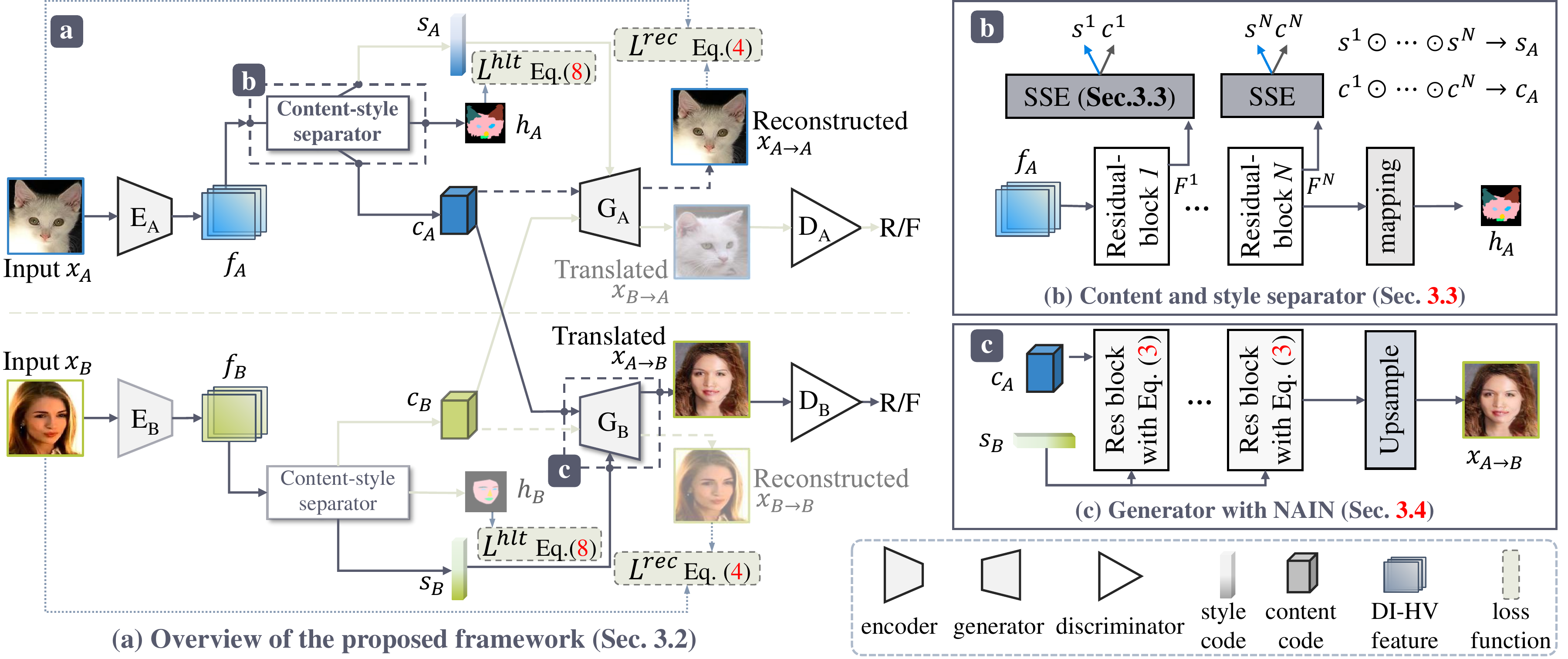} 
	\end{center}
	\vspace{-0.2cm}
	\caption{System architecture. (a) Overview of the proposed \proposed~framework for UIT. Take cat ($x_A$) to human ($x_{A\to B}$) as a example, $x_A$ is first mapped into feature $f_A$ through encoder $E_A$. Then we separate content and style based on the correlation between the feature and \DIT~task, \ie, facial parsing. Next we synthesize the image using a generator with NAIN, which combines the content code $c_A$ and the style code $s_B$ together to generate the $x_{A\to B}$. (b) The structure of content-style separator. Note that $\odot$ is channel-wise concatenation. (c) The structure of generator with NAIN.}
	\label{fig:overview}
	\vspace{-0.4cm}
\end{figure}

Fig.~\ref{fig:overview} (a) demonstrates the overview of our proposed method. In addition to the basic encoders $\{E_A, E_B\}$ for \DIT, the generators $\{G_A, G_B\}$ and the discriminators $\{D_A, D_B\}$, our method also learns the correlation-based feature separation functions $\Pi_{F_A\to \C, \SA, \H}$ and $\Pi_{F_B\to \C, \SB, \H} $, which 1) separates the \DIT~ features into the domain-invariant content space $\C$ and domain-specific style space $\mathcal{S}$, and 2) maps the feature space into the \DIT~task space $\H$. For the mapping of $\XA \mapsto \XB$, the source image $x_A \in \XA$ is first encoded into the \DIT~ feature: $f_A = E_A(x_A)$. Next, the content code $c_A$ and style code $s_A$ are separated through $c_A, s_A = \Pi_{F_A\to \C, \SA}(f_A)$. Then the generator $G_B$ takes the content code $c_A$ and the style code $s_B$, which is sampled from $\SB$, for translating the output image $x_{A\to B}$. In summary, for the translation of $\XA \mapsto \XB$, our method synthesizes the output by

\begin{equation}
	x_{A\to B} = G_B(\Pi_{F_A\to \C_A}(E_A(x_A)), s_B).
\end{equation}

Similarly, the translation of $\XB \mapsto \XA$ is achieved by

\begin{equation}
	x_{B\to A} = G_A(\Pi_{F_B\to \C_B}(E_B(x_B)), s_A).
\end{equation}

\subsection{Learning to separate content and style} \label{sec:main_module}

The critical part is to extract content code and style code from the feature space with $\Pi_{F_A\to \C, \SA}$ and $\Pi_{F_B\to \C, \SB}$, and to map features to \DIT~ task space.
As illustrated in Fig.~\ref{fig:overview} (b), the content-style separator consists of several residual blocks and a \DIT~task mapping function, where each of them is connected with a Squeeze-Selection-and-Excitation (SSE) module.

\noindent\textbf{Squeeze-Selection-and-Excitation for Content-style Separation.} 
The SSE module is inspired by the SE-Networks~\cite{V:hu2018squeeze}, which learns weights for different feature channels. The weights describe how important the channels contribute to the final task. In this paper, we take the weight as a descriptor of the correspondence between the feature and \DIT~ task. We denote each channel of the feature as a \textit{unit}. Units with high correspondence scores are taken as content code, and the rest of units are taken as style code. In detail, 1) we \textit{squeeze} the units in feature $f$ into channel descriptors via an adaptive average pooling layer. Following ~\cite{V:hu2018squeeze}, we compute the channel weights by learning a non-linear mapping. The weights are taken as the correlation score of each unit. 2) We \textit{select} the units with high correlation scores as content code $c$, and the rest of units as style code $s$. 3) The correlation scores then \textit{excite} the units in the feature $f$ for \DIT~ task. 
% The details of the SSE algorithm can be found in the supplementary materials.
The input of the SSE module is the feature $F_k$, which is encoded from the input image. Then this module separates the content code and style code from $F_k$. The detailed steps are shown in the supplement materials (Algorithm~\ref{algor:SSE}).
% the supplementary materials.

\noindent\textbf{Flexible mappings for \DIT~ tasks.}
We map the learned features to the \DIT~tasks using supervised learning. The decoder is flexible for different \DIT~tasks. For example, if the image is from domain-invariant dataset (\eg MNIST), \DIT~task decoder is implemented as a fully connected layer for classification; if the scene semantic segmentation map is available, the high-level task mapping is constructed with a fully convolutional network.

\subsection{Image generation with NAIN}

In our experiments, we found that the water-drop effects often appeared in the synthesized image. 
The water-drop effects result in the quality degradation of the generated images. To remove such effects, we propose a normalized adaptive instance normalization (NAIN) block to improve the quality of the generated image.

As reported in StyleGAN2~\cite{G:StyleGAN2}, the effects are mainly caused by the abnormal bias in Adaptive Instance Normalization (AdaIN)~\cite{R:huang2017AdaIN}, which is used in  generator for fusing content code and style code.
Different from Karras~\etal~\cite{G:StyleGAN2}, where they added a `demodulation' operation, we propose a solution within a unified layer, \ie, Normalized AdaIN (NAIN), to restrain the intermediate parameters in AdaIN.
To be more specific, we normalize the parameters of AdaIN before assigning them to the instance normalization layer. The NAIN is defined as

\begin{equation} \label{eq:nain}
	\begin{aligned}
		\text{NAIN}(z, \gamma, \beta) = \sigma(\gamma) (\frac{z - \mu(z)}{\sigma(z)}) + \sigma(\beta),
	\end{aligned}
\end{equation}
where $\sigma(z) = 1 / (1 + \exp(-z))$, $z$ is the activation of the previous convolutional layer, $\mu$ and $\sigma$ are channel-wise mean and standard deviation, $\gamma$ and $\beta$ are parameters generated by the multi-layer perceptron, which takes style code as input. NAIN prevents the intermediate variable $z$ from too large or too small values.

\subsection{Loss functions}
There are four loss functions in our model: image reconstruction loss $\LL^{\mathrm{rec}}$, adversarial loss $\LL^{\mathrm{adv}}$, perceptual loss $\LL^{\mathrm{perc}}$ and \DIT~ task loss $\LL^{\mathrm{hlt}}$.

Once the input image is reconstructed through auto-encoder with supervised learning, we compute the reconstruction loss as:
\begin{equation} \label{eq:rec_loss}
	\LL^{\mathrm{rec}} = |x_A - x_{A\to A}|_1 + |x_B - x_{B\to B}|_1.
\end{equation}

To stabilize and accelerate the the GAN training, we use the LSGAN objective proposed by Mao \etal~\cite{G:mao2017least}. The least-squares adversarial losses~\cite{G:mao2017least} are defined as follows: 
\begin{equation} \label{eq:adv_loss}
	\LL^{\mathrm{adv}}_{\XA} = (1-D_A(x_A))^2 + (D_A(x_{B\to A}))^2,~~
	\LL^{\mathrm{adv}}_{\XB} = (1-D_B(x_B))^2 + (D_A(x_{A\to B}))^2.
\end{equation}
The total adversarial loss is $\LL^{\mathrm{adv}} = \LL^{\mathrm{adv}}_{\XA} + \LL^{\mathrm{adv}}_{\XB}$.

We also add domain-invariant perceptual loss to make the results more realistic. The perceptual loss, often computed as a distance in the VGG~\cite{B:simonyan2014very} feature space between the output and the reference image, accelerates training on high-resolution datasets.
The perceptual loss is defined as below:
\begin{equation} \label{eq:vgg_loss}
	\LL^{\mathrm{perc}} = |\Psi(x_A) - \Psi(x_{A\to B})|_1 + |\Psi(x_B) - \Psi(x_{B\to A})|_1,
\end{equation}
where $\Psi(x)$ is the high-level feature of input image $x$ in VGG. We follow the settings of the perceptual loss from MUNIT~\cite{I:MUNIT} in our experiments.

For \DIT~task, if the semantic segmentation or classification is applied, we use cross entropy as the loss function:
\begin{equation}\label{equ:loss_semantic}
	\LL^{\mathrm{hlt}} = \sum_{j=1}^{K}\sum_{i=1}^{M}\left(-y_{ji}\log{\hat{y}_{ji}} - \left(1-y_{ji}\right)\log{\left(1-\hat{y}_{ji}\right)}\right),
\end{equation}
where $K$ is the batch size and $M$ is the summation over classes, $\hat{y}$ is the prediction and the ground truth label is $y$.

\textit{Total loss.}
By using the GAN scheme, we jointly train the encoders $E$, content-style separator $\Pi$, decoders $G$ and discriminators $D$ to optimize the weighted sum of the different loss terms:
\begin{equation} \label{eq:objective_func}
		\min_{E, \Pi, G}\max_{D}(E, \Pi, G, D) =  \LL^{\mathrm{adv}} + \lambda_1 * \LL^{\mathrm{rec}}
		 + \lambda_2 * \LL^{\mathrm{perc}} + \lambda_3 * \LL^{\mathrm{hlt}},
\end{equation}
where $\lambda_1, \lambda_2, \lambda_3$ are hyper-parameters for adjusting the weights of each loss function.

%-------------------------------------------------------------------------
\section{Experiments}
\label{sec:experiments}

\textbf{Datasets.}
We conducted extensive experiments on four datasets.

% \noindent\textit{Cat$\leftrightarrows$Human} is a dataset for translation between cat face and human face. Face parsing is used for \DIT~task.

% \noindent\textit{Man$\leftrightarrows$Woman} is similar to Cat$\leftrightarrows$Human. Images are sampled from CelebA~\cite{D:liu2018large}.

% \noindent\textit{Colored-MNIST}. We follow \cite{I:gonzalez2018image} and generate images with colored digits on black background or white digits on colored background. 
% Number classification is used for the \DIT~task.

% \noindent\textit{CG$\leftrightarrows$Real} dataset is proposed by \cite{I:bi2019deepCG2Real} and is used for translation between synthetic indoor images and real indoor scenes. 
% Semantic scene parsing is used as the \DIT~task.

\noindent \textit{Cat$\leftrightarrows$Human} is obtained from the Kaggle~\footnote{\url{https://www.kaggle.com/crawford/cat-dataset}} and CelebA~\cite{D:liu2018large}, which are publicly available. For each population of images in different domains, we collected 6000 images (5000 for training and 1000 for testing). Semantic segmentation is taken as the \DIT. Specifically, we generated the pseudo semantic segmentation mask based on the cat/human facial landmark location.

\noindent \textit{Man$\leftrightarrows$Woman} dataset is generated by randomly sampling 10000 images (9000 for training and 1000 for testing) from CelebA~\cite{D:liu2018large} for each domain. The semantic segmentation mask is generated in the same way as Cat$\leftrightarrows$Human.
	
\noindent \textit{Colored-MNIST} dataset is an extension of MNIST dataset~\cite{A:lecun1998mnist}. We followed \cite{I:gonzalez2018image} and generated images with colored digits on black background or white digits on colored background. The image classification is used for the \DIT.
	
\noindent \textit{CG$\leftrightarrows$Real} dataset is proposed by \cite{I:bi2019deepCG2Real}, which translates between synthetic indoor images and real indoor scenes. 
Semantic scene parsing is used as the \DIT, where we generate the pseudo semantic segmentation mask by HRNet~\cite{A:WangSCJDZLMTWLX19}. The HRNet is trained on ADE20K~\cite{A:zhou2017ADE20K}.
	
We have also tested our methods on other \uit~ tasks provided by \cite{I:MUNIT}. These tasks include \textit{Cat$\leftrightarrows$Dog} (face parsing is used as \DIT), and \textit{Summer$\leftrightarrows$Winter} (semantic segmentation~\cite{A:WangSCJDZLMTWLX19} is used as \DIT). 
	
% \noindent \textit{Cat$\leftrightarrows$Dog} dataset is provided by \cite{I:MUNIT}. Here we use facial parsing as \DIT, in which labels are pre-computed by xx.

% \noindent \textit{Summer$\leftrightarrows$Winter} dataset is provided by \cite{I:MUNIT}. Here we use semantic segmentation~\cite{A:WangSCJDZLMTWLX19} as \DIT. The other settings are same as CG$\leftrightarrows$Real.

% \noindent \textit{Night$\leftrightarrows$Day} dataset is provided by \cite{I:MUNIT}. Here we use semantic segmentation as \DIT. The other settings are same as CG$\leftrightarrows$Real.

% TODO:
\noindent\textbf{Evaluation metrics.}
We evaluated both the realism and diversity of the generated images for different methods. We adopted the Fr{\'e}chet Inception Distance~\cite{M:heusel2017FID} (FID) to measure visual quality, which is a general image quality assessment method. To measure diversity, we used the Learned Perceptual Image Patch Similarity (LPIPS) to calculate the diversity among images, similar to~\cite{M:zhang2018LPIPS}. Both metrics use InceptionV3~\cite{B:szegedy2016rethinking} as the backbone, which is pretrained on ImageNet.

\noindent\textbf{Implementation details.}
Inspired by MUNIT~\cite{I:MUNIT}, we constructed the \proposed~ using a similar decoder and discriminator.
Images of Cat$\leftrightarrows$Human, Man$\leftrightarrows$Woman, and CG$\leftrightarrows$Real dataset are randomly cropped and resized to $256 \times 256$ before feeding into the model. In Algorithm~\ref{algor:SSE}, we set select ratio $r = 0.25$. We used 4 residual blocks for the content-style separator. For \DIT mapping, we applied 2 convolutional blocks with an instance normalization layer for semantic estimation. To optimize the training complexity and performance, we empirically set multi-scale to 3 in the discriminator. We set the iteration to 1e8 in the experiments. 
Since images in Colored-MNIST are in low resolution, we resized them to $64 \times 64$ and use 2 residual blocks for content-style separator. A two-layer MLP is adopted for classification. Other hyper-parameters are set the same as other datasets. We empirically set $\lambda_1 = 5, \lambda_2 = 0.5, \lambda_3 = 1$ for the objective in \eq~\eqref{eq:objective_func}.

\subsection{Interpretation of the learned representations}

The proposed framework is able to interpret the translation results. Inspired by the image classification task~\cite{V:zeiler2014visualizing} and recent GAN dissection~\cite{V:bau2018GANdissection}, we find the feature maps, which are from different channels of the our learned content code, represent different semantic information of the input image. Based on this observation, we further manipulate the translations with semantic meanings through operations in feature space.

% In classifier networks, the type of information explicitly represented changes from layer to layer~\cite{V:zeiler2014visualizing}. A similar phenomenon in a GAN is also found by GAN dissection~\cite{V:bau2018GANdissection}. We extend them and find the similar phenomenon in our framework, which further enables us to manipulate the translations with semantic meaning.

\begin{figure} [htbp]
	\begin{minipage}[c]{0.62\textwidth}
		\includegraphics[width=\textwidth]{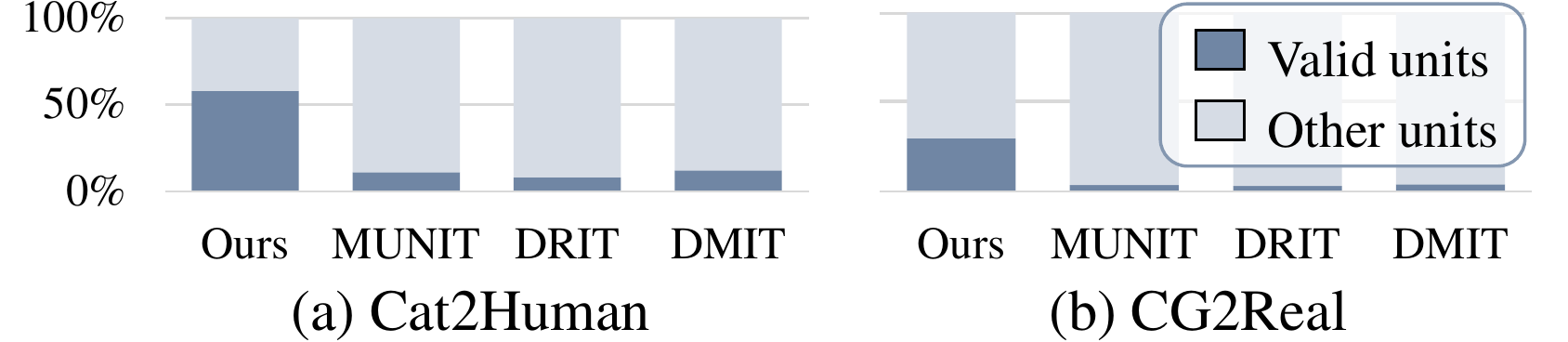}
	\end{minipage}\hfill
	\begin{minipage}[c]{0.35\textwidth}
		\caption{
			Comparison of the number of valid unit in content code between different methods.
		} \label{fig:valid_units_cmp}
	\end{minipage}
	\vspace{-0.8cm}
\end{figure}

We extract the content codes (64 units) of the input image and find the valid units by calculating the IoUs between the unit and semantic segmentation. Intuitively, if a unit has high overlap with a specific semantic map (\eg, facial region), it will be a valid unit for the facial semantics. Mathematically, the valid unit is defined as follows:
\begin{equation} \label{equ:semantic_units}
	\begin{aligned}
		U_{\mathrm{valid}}^m = \begin{cases}
			1,  & \text{if } \mathtt{IoU}(U^*, M) > \alpha; \\
			0,  & \text{otherwise.}
		\end{cases} 
	\end{aligned}
\end{equation}
where $U_{i, j}^* = 1$, if $U_{i, j} > \beta$; otherwise, we set $U_{i, j}$ to 0. $\alpha$ and $\beta$ are thresholds for IoU and activated units $U^*$. 
$M$ is the semantic map, which is used as ground-truth for \DIT~task.
We set $\alpha=0.18$, $\beta=0.4$ in the experiment.
Fig.~\ref{fig:valid_units_cmp} demonstrates our method is able to extract more semantically valid units in the content code across different image-to-image translation tasks. In each task, we tested 100 images on visualizing the relationship between the semantic masks and the content codes. 

\subsection{Semantic area manipulation}
% \noindent\textbf{Manipulating semantic area for translation.}
The proposed framework is capable of interactive manipulation of the semantic areas in the translated images. Here, we conducted two experiments: replacing the semantic area and ablating the artifacts.

\begin{figure} [htbp]
	% \vspace{-0.2cm}
	\begin{center}
		\includegraphics[width=0.9\linewidth]{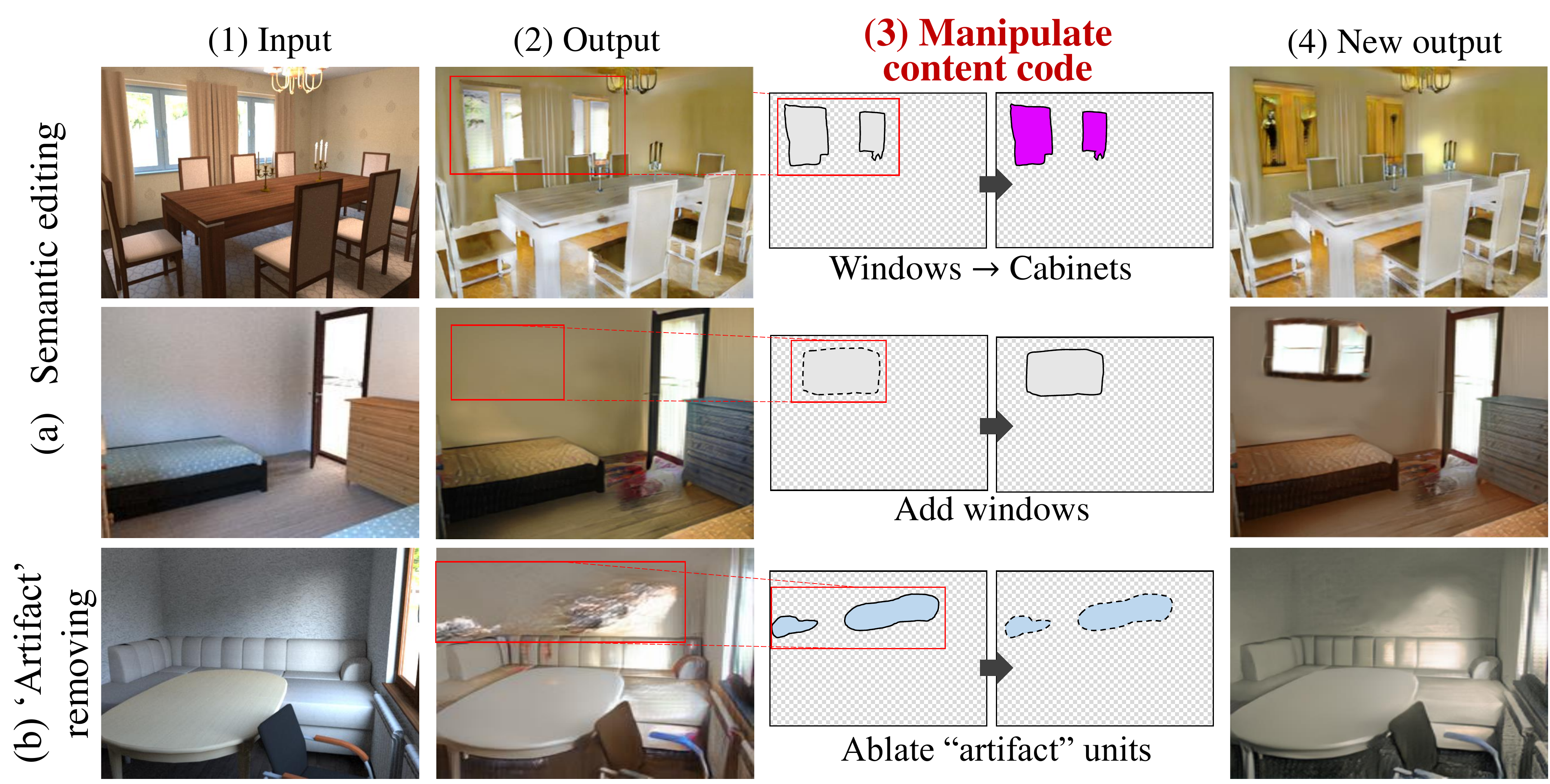} 
	\end{center}
	\vspace{-0.4cm}
	\caption{Editing results on the CG$\leftrightarrows$Real dataset. The proposed method is capable of (a) editing the semantic regions, such as replacing the windows to cabinets, adding windows on the wall, and (b) removing the `artifact' units.
	}
	\label{fig:manipute_cg2real}
% 	\vspace{-0.2cm}
\end{figure}

We first show the results of editing the semantic area. Take the first row of Fig.~\ref{fig:manipute_cg2real} as an example, 
1) we locate the IDs of the `windows' units through \eq~\eqref{equ:semantic_units}. 
2) We masked the window's region and replaced the activated area (as shown in the third column) with $\min (U_w)$, where $U_w$ is the units for windows. 
3) Similar to step-1), we located the `cabinets' units' channel ids. 
4) We obtained the manipulated units by filling $\max (U_c)$ ($U_c$ is the units for the cabinets) on the `cabinets' units' masked areas. 
5) Conditioned by the edited units, the decoder generates new output, in which the area of windows is replaced with cabinets.

We find the proposed framework can also be used to remove artifacts. As shown in the last row of Fig.~\ref{fig:manipute_cg2real}. By manually masking the artifacts area, the `artifacts' units can be found through \eq~\eqref{equ:semantic_units}. Ablating such units improves the quality of translated images.

\begin{table}[htbp]
	\begin{center}
		\setlength{\tabcolsep}{1mm}
		\renewcommand\arraystretch{1.2}
		\caption{Quantitative Results. We use FID (the lower the better) to measure the quality and LPIPS distance (the higher the better) for the diversity of the translated images.}
		\label{tab:numerical_comparisons_sota}
		\footnotesize
		\begin{tabular}{l cc cc cc cc cc cc}
			\toprule[1.2pt]
			
			Tasks & \multicolumn{ 2}{c}{DCLGAN~\cite{I:han2021dcl}} & \multicolumn{ 2}{c}{DSMAP~\cite{I:chang2020dsmap}} & \multicolumn{ 2}{c}{MUNIT~\cite{I:MUNIT}} & \multicolumn{ 2}{c}{DRIT~\cite{I:DRIT}} & \multicolumn{ 2}{c}{DMIT~\cite{I:yu2019DMIT}} & \multicolumn{ 2}{c}{Ours} \\
			& FID & LPIPS & FID & LPIPS & FID & LPIPS & FID & LPIPS & FID & LPIPS & FID & LPIPS \\
			\hline
			\specialrule{0em}{1pt}{1pt}
			
			\multirow{2}{1.5cm}{Cat$\leftrightarrows$Human} & 83.00 & 0.331 & 103.65 & 0.315 &  89.01 & 0.362 & 123.26 & 0.216 & 225.49 & 0.285 & \textbf{72.53} & \textbf{0.469} \\
			&                               77.22 & 0.501 & 160.10 & 0.417 &  90.50 & 0.315 & 166.18 & 0.259 & 228.28 & 0.240 & \textbf{68.66} & \textbf{0.577} \\
			\rowcolor{mygray}
			&                               59.65 & 0.369 &  57.34 & 0.369 &  66.61 & 0.264 &  65.86 & 0.152 &  71.23 & 0.311 & \textbf{57.17} & \textbf{0.398} \\
			\rowcolor{mygray}
			\multirow{-2}{1.5cm}{Man$\leftrightarrows$Woman}  & 54.70 & 0.394 &  64.44 & 0.346 &  \textbf{48.41} & 0.328 &  98.07 & 0.161 &  94.98 & 0.319 & 65.12 & \textbf{0.419} \\
			&                               59.99 & 0.433 &  \textbf{53.56} & 0.421 &  56.37 & 0.336 & 69.90 & 0.403 &  65.50 & \textbf{0.460} & 58.97 & 0.445 \\
			\multirow{-2}{1.5cm}{Colored-MNIST} & 31.06 & 0.216 &  35.29 & 0.401 &  30.25 & 0.205 & 70.53 & 0.265 &  37.60 & 0.180 & \textbf{28.63} & \textbf{0.269} \\
			\rowcolor{mygray}
			&                                75.95 & 0.391 &  83.31 & 0.299 &  74.61 & 0.340 &  79.98 & 0.208 &  73.11 & 0.243 & \textbf{70.07} & \textbf{0.503} \\
			\rowcolor{mygray}
			\multirow{-2}{1.5cm}{CG$\leftrightarrows$Real} & 70.07 & 0.476 &  76.07 & \textbf{0.515} &  79.53 & 0.259 &  85.47 & 0.211 &  72.79 & 0.227 & \textbf{68.71} & 0.449 \\
			\hline
			Average                & 63.95 & 0.389 &  79.22 & 0.385 &  66.91 & 0.301 & 94.91 & 0.247 & 108.63 & 0.283 &  \textbf{61.27} & \textbf{0.441} \\
			
			\bottomrule[1.2pt]
		\end{tabular}
	\end{center}
	\vspace{-0.8cm}
\end{table}

\subsection{Performance comparison}
\begin{figure} [htbp]
	\begin{center}
		\includegraphics[width=\linewidth]{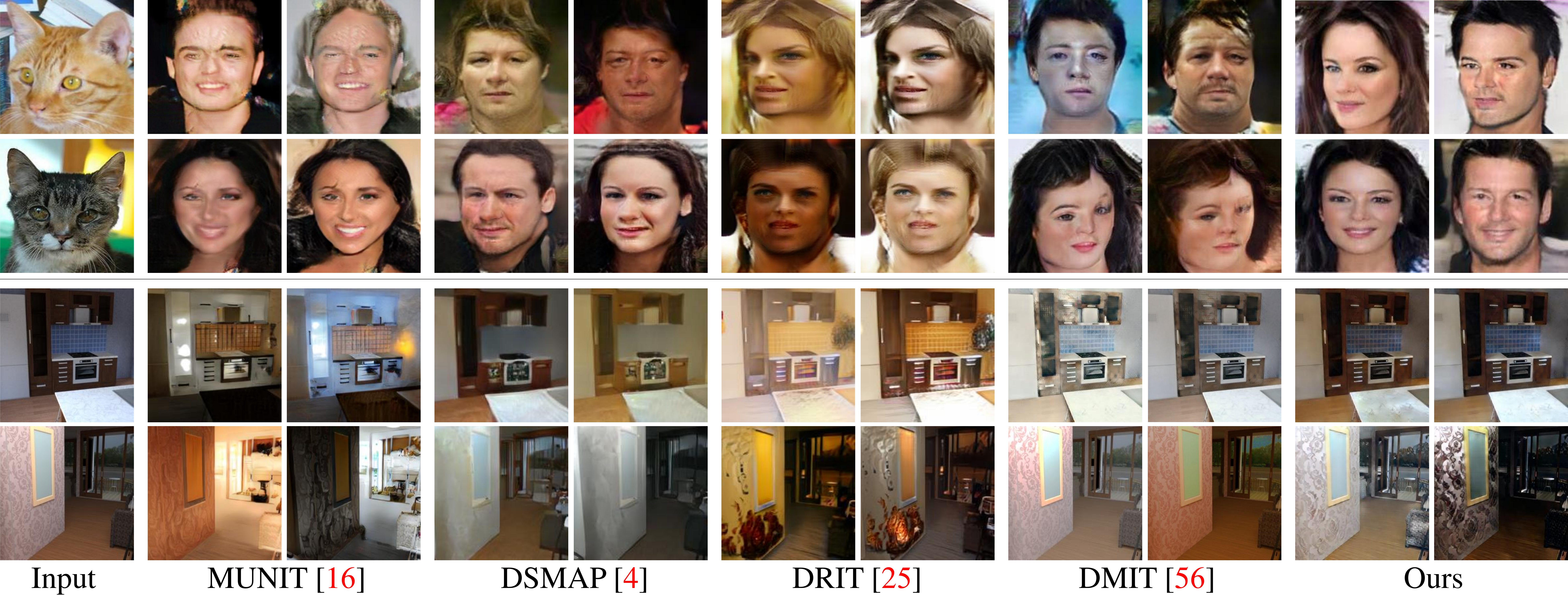} 
	\end{center}
	\vspace{-0.4cm}
	\caption{Qualitative comparison on Cat to Human (Top) and on the CG to Real (Bottom).}
	\label{fig:sota_cmp}
	\vspace{-0.4cm}
\end{figure}

We qualitatively and quantitatively compared our method with the four state-of-the-art UIT methods: DCLGAN~\cite{I:han2021dcl}, DSMAP~\cite{I:chang2020dsmap}, MUINT~\cite{I:MUNIT}, DRIT~\cite{I:DRIT} and DMIT~\cite{I:yu2019DMIT}. We train these four baselines on our datasets with their publicly available implementations.
As shown in Table~\ref{tab:numerical_comparisons_sota}, the proposed \proposed~framework outperforms the other methods on four datasets in terms of visual quality (\ie, FID) and diversity (\ie, LPIPS). 
In the first two rows of Fig.~\ref{fig:sota_cmp}, we provide a visual comparison on Cat$\leftrightarrows$Human task. The results show that the geometry of the input image and translated image are more consistent than other methods. The last two rows of Fig.~\ref{fig:sota_cmp} compare the performance of different methods on CG$\leftrightarrows$Real dataset. Visual results show that our method generates more realistic textures (first row), diverse illuminations (last row).
% TODO: add these images to the main file? 
More visual results on these two tasks, Colored-MNIST and Man$\leftrightarrows$Woman are provided in the supplementary materials.

\begin{figure} [htbp]
	\begin{center}
		\includegraphics[width=\linewidth]{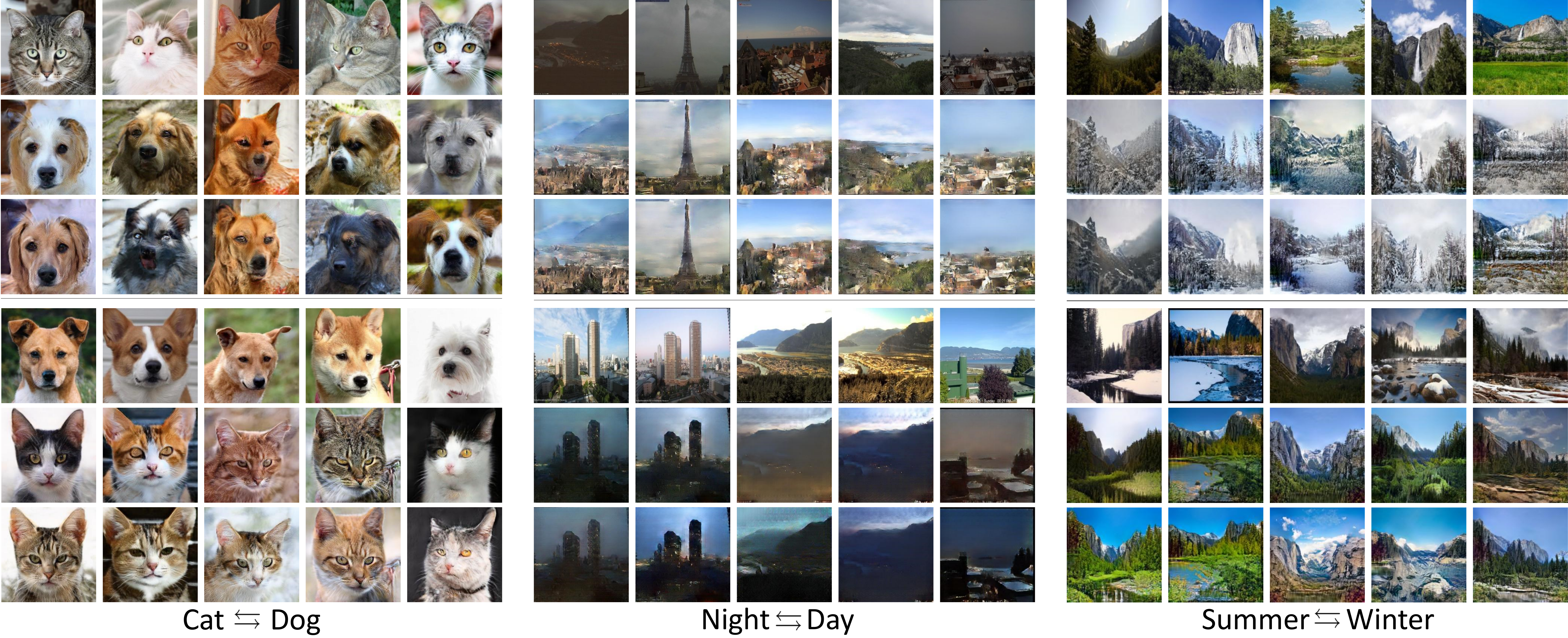} 
	\end{center}
	\vspace{-0.4cm}
	\caption{Visual results of our method on other three \uit~tasks. For each task, we provide bidirectional translations, which are separated by a solid line. In each translation, we provide 5 inputs (first row) and the corresponding multimodal results (the second and third row).}
	\label{fig:sota_cmp_supp}
	\vspace{-0.4cm}
\end{figure}

\subsection{Ablation study}

To verify the effectiveness of NAIN and SSE, we perform rigorous ablation studies by changing the different parts of our method.
We analyze the SSE module by 1) replacing the proposed encoder with two encoders for content code and style code, respectively, and the framework is similar to MUNIT~\cite{I:MUNIT}; 2) removing the SSE module (w/o SSE), and extract the content code and style code statically by setting the first quarter of the units as content, and the remaining part as style; 3) replacing NAIN with the original AdaIN (w/o NAIN). 
Fig.~\ref{fig:ablation} and Table ~\ref{tab:ablation} consistently show the final version of \proposed~get the best performance both qualitatively and quantitatively. Our full model can improve the content consistency (vs. two encoders), improve the resulting quality (vs. w/o SSE) and remove the water-drop artifact effectively (vs. w/o NAIN).

\begin{table}[ht]
\centering
\begin{minipage}[c]{0.51\textwidth}
		\centering
		\includegraphics[width=\linewidth]{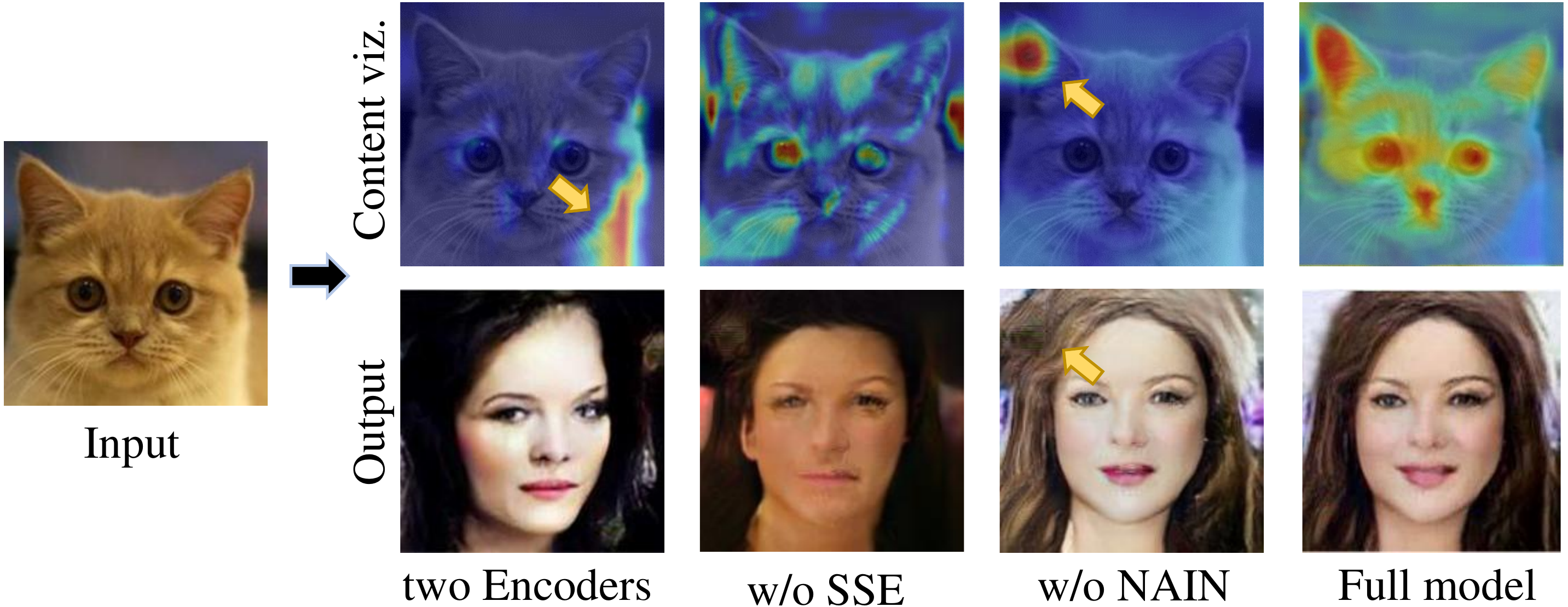}
		\vspace{-7mm}
		\captionof{figure}{Visual comparisons of ablation study.  
		}\label{fig:ablation}
	\end{minipage}\hfill
	\begin{minipage}[c]{0.40\textwidth}
		\footnotesize
		\caption{Numerical results of ablation study.}%
		\label{tab:ablation}
		\begin{tabular}{ccc}
			\toprule[1.2pt]
			
			& FID $\downarrow$ & LPIPS $\uparrow$ \\
			\hline
			\specialrule{0em}{1pt}{1pt}
			
			Two Encoders  & 119.87 & 0.339\\
			\rowcolor{mygray}
			w/o SSE       & 121.31 & 0.284\\
			w/o NAIN      & 78.92  & 0.402 \\
			\rowcolor{mygray}
			Ours          & \textbf{72.53}  & \textbf{0.449}\\
			
			\bottomrule[1.2pt]
		\end{tabular}
	\end{minipage}
	\vspace{-0.6cm}
\end{table}

\section{Conclusion}

In this paper, we proposed a \proposed~framework for unsupervised image-to-image translation. We introduced the domain-invariant high-level vision task for human-perceptive translation. We used an SSE module to separate content code and style code on the basis of relevance between the features and the \DIT~within one unified encoder. The proposed method enables us to interpret and manipulate the translated image. Experimental results on four translation data sets demonstrate that our method outperforms the state-of-the-art methods in terms of visual quality and diversity. This work opens the pathway towards interpreting and manipulating the results in image translation tasks.

\bibliography{egbib}

\newpage
\appendix

{\centering\huge Supplementary Material for \\ \textit{Separating Content and Style for Unsupervised Image-to-Image Translation}}

\section{Algorithm of Squeeze-Selection-and-Excitation for Content-style Separation}

The input of the SSE module is the feature $F_k$, which is encoded from the input image. Then this module is aim to separate the content code and style code from $F_k$. The detailed algorithm is shown in Algorithm~\ref{algor:SSE}.

\begin{algorithm}[htbp] % Move it to appendix
	\caption{Squeeze-Selection-and-Excitation (SSE) for content-style separation}  
	\label{algor:SSE}
	\KwIn{The \textit{k-th} residual block's feature map $F^k$, selection ratio $r$}
	\KwOut{The corresponding content code $c^k$ and style code $s^k$}
	
	$a \gets \mathtt{channels\_of}(F^k)$ \Comment $a$ is the number of channels
	
	$T_1 \gets \mathtt{adaptive\_average\_pooling}(F^k)$ \Comment Squeeze
	
	$T_2 \gets \Phi_1(T_1)$ \Comment $\Phi_1$ is a Multi Layer Perceptron (MLP) 
	
	$I \gets \mathtt{index\_of\_descend\_sort}(T_2, 1)$ \Comment $I$ is the channel index
	
	$I_c \gets \text{first } \lfloor a * r \rfloor \text{ elements in } I$ \Comment High correspondence are selected as content
	
	$I_s \gets I \setminus I_c $	\Comment The rest are taken as style
	
	$c^k, t \gets F^k[I_c], F^k[I_s]$
	
	$s^k \gets \Phi_2 (t)$ \Comment $\Phi_2$ is an encoder for compressing the style code
	
	$F_k$ = $F_k \cdot T_2$ 	\Comment Excitation
	
	\textbf{return} $c^k$, $s^k$
\end{algorithm} 

% \section{Details in Dataset.}

% Details of the dataset are as follow.
% \begin{itemize}
% 	\item \textit{Cat$\leftrightarrows$Human} is obtained from the publicly available Kaggle~\footnote{\url{https://www.kaggle.com/crawford/cat-dataset}} and CelebA~\cite{D:liu2018large}. For each population of images in different domain, we collect 6000 images (5000 for training and 1000 for testing).
% 	Semantic segmentation is taken as the high-level domain-invariant task. Specifically, we generate the pseudo semantic segmentation mask on the basis of the cat/human facial landmark location.
% 	\item \textit{Man$\leftrightarrows$Woman} dataset is randomly sampled 10000 images (9000 for training and 1000 for testing) from CelebA~\cite{D:liu2018large} for each domain.
% 	The semantic segmentation mask is generated as the same way as Cat$\leftrightarrows$Human.
% 	\item \textit{Colored-MNIST} dataset is an extension of MNIST dataset~\cite{A:lecun1998mnist}. We follow \cite{I:gonzalez2018image} and generate images with colored digits on black background or white digits on colored background. 
% 	The image classification is used for the high-level domain-invariant task.
% 	\item \textit{CG$\leftrightarrows$Real} dataset is proposed by \cite{I:bi2019deepCG2Real} and is used for translation between synthetic indoor images and real indoor scenes. 
% 	Semantic scene parsing is used as the high-level task, in which we generate the semantic segmentation mask through HRNet~\cite{A:WangSCJDZLMTWLX19}, which is trained on ADE20K~\cite{A:zhou2017ADE20K}.
% \end{itemize}

\section{More Details About the Valid Units for Interpretation.}

\begin{figure} [htbp]
	\includegraphics[width=\textwidth]{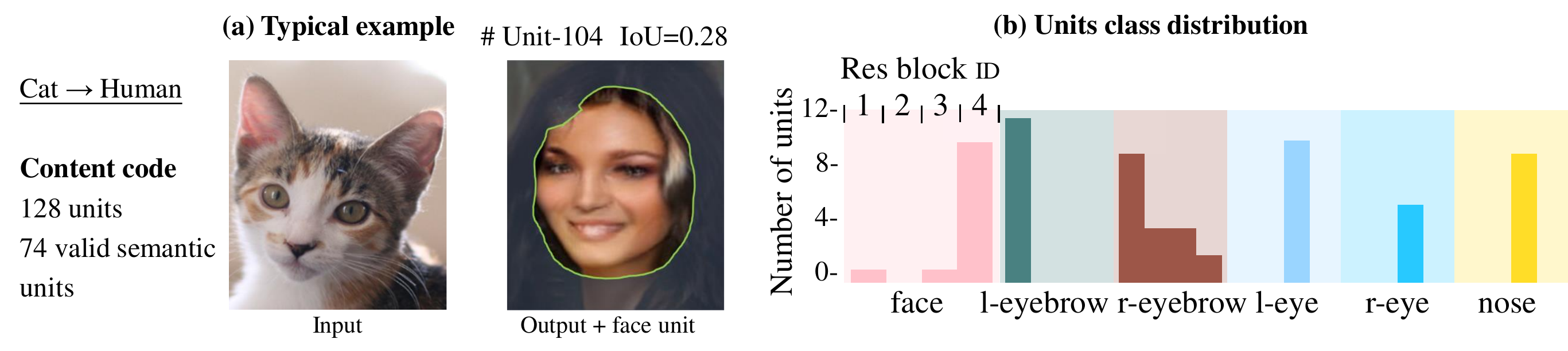}
	\caption{Interpretation of the feature maps in content code, which is learned on Cat$\leftrightarrows$Human task. (a) shows the valid face unit focuses on the correct region of the output image. (b) illustrates that all the semantic parts have multiple valid units in content code.
	} \label{fig:viz}
	\vspace{-0.4cm}
\end{figure}

In detail, Fig.~\ref{fig:viz} (a) shows an example of the activated face features on the generated human image. We find the valid face unit reflects the shared attention face area between the input and output images. There are 6 classes for different semantic parts in Cat$\leftrightarrows$Human dataset: face, left/right eye brow, left/right eye and nose. Fig.~\ref{fig:viz} (b) shows the units class distribution. Different bins in the same semantic histogram denote different content codes yield from four different residual blocks in encoder. There is one peak value in the face histogram, which indicates that the last residual block yield content code that contains most of the facial features. The residual block ID with larger value extracts more global features~\cite{B:he2016deep}, it can be interpreted that the facial units gather global information. The other semantic units are illustrated in the other histograms with different colors. The semantic units between different domains exist at similar position of the \proposed. This is because the semantic segmentation is domain-invariant, and the \proposed~can learn the similar semantic parts within the same units in the encoder.

\begin{figure} [htbp]
	\begin{center}
		\includegraphics[width=\linewidth]{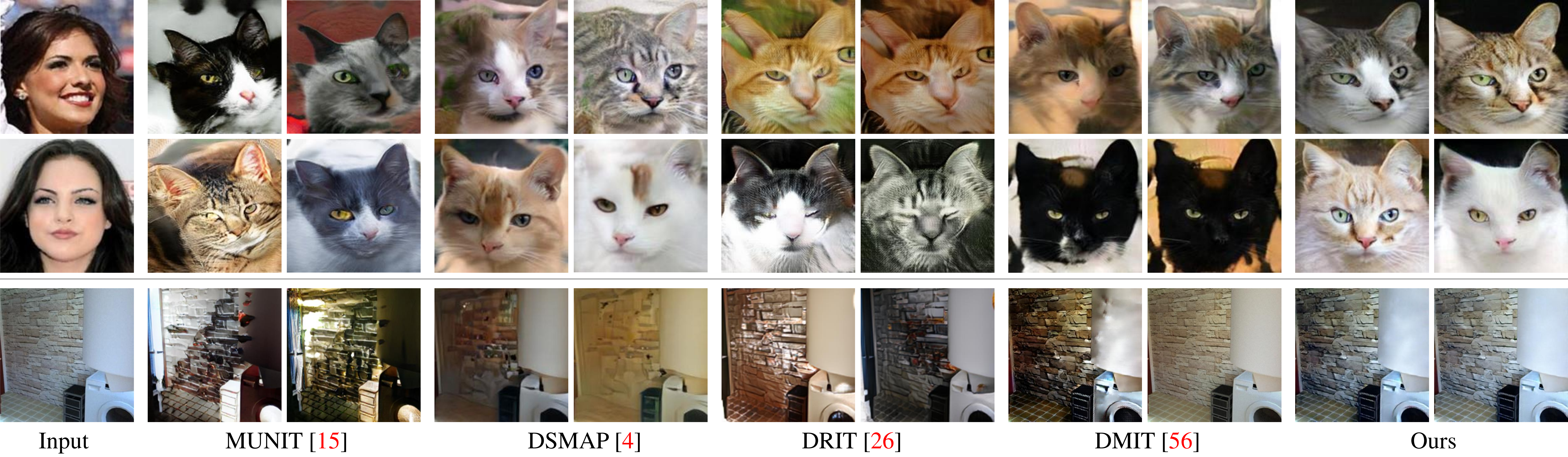} 
	\end{center}
	\vspace{-0.4cm}
	\caption{More qualitative comparison on Human to Cat task (Top) and on the CG to Real task (Bottom).}
	\label{fig:sota_cmp_supp}
	\vspace{-0.4cm}
\end{figure}

\begin{figure} [htbp]
	\begin{minipage}[c]{0.72\textwidth}
		\includegraphics[width=\linewidth]{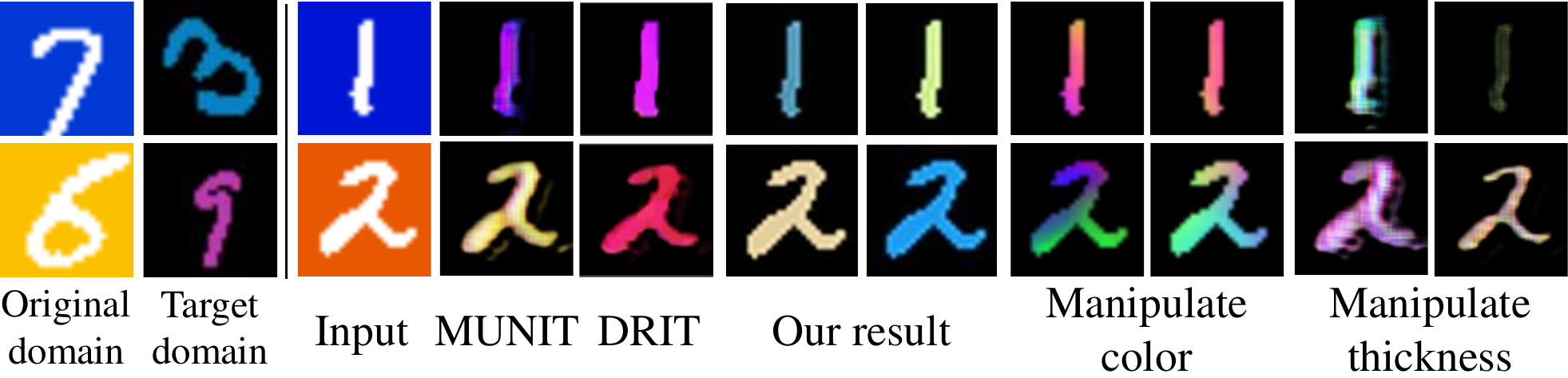} 
		\vspace{1mm}
		\includegraphics[width=\linewidth]{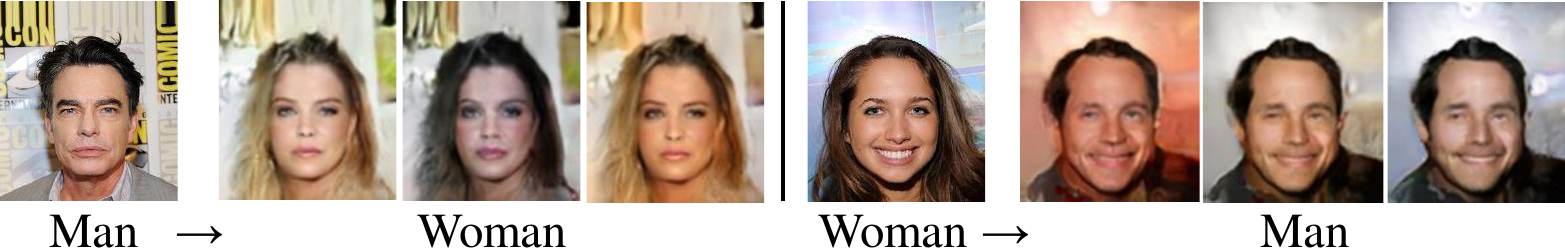} 
	\end{minipage}\hfill
	\begin{minipage}[c]{0.25\textwidth}
		\caption{
			Comparison and manipulation results on Colored-MNIST dataset(top) and multimodal results on Man$\leftrightarrows$Woman Task (bottom).
		} \label{fig:more_results}
	\end{minipage}
	\vspace{-0.4cm}
\end{figure}

\section{More Visual Comparison Results}

We show more visual results in Fig.~\ref{fig:sota_cmp_supp} on Human to cat task and CG to Real task.

Take the `Colored MNIST' as an example, which in shown in top row of Fig.~\ref{fig:more_results}. To manipulate the styles of the generated images, we add random noise to the intermediate style code by $s_1 = s_1 + e$, where $e \sim U(0, 1)$, and $s_1$ is defined in Fig.~\ref{fig:overview}. we find more than one colors appear on the digits number, which shows more various and possible results than target domain. Furthermore, we can manipulate these units by adding image morphology operation (\eg, dilation and erosion). It can be observed that the digital number become thicker and thinner.

\end{document}